\documentclass{article}
\usepackage{paperstyle,times}
\usepackage{url}
\usepackage{graphicx}
\usepackage{amsmath,amsfonts,booktabs}
\usepackage{hyperref}
\setcitestyle{citesep={,},yysep={,}}
\hypersetup{colorlinks=false,pdfborder={0 0 1},citebordercolor={0 1 0}}
\makeatletter
\long\def\@makecaption#1#2{%
  \vskip\abovecaptionskip
  \sbox\@tempboxa{#1. #2}%
  \ifdim \wd\@tempboxa >\hsize
    #1. #2\par
  \else
    \global \@minipagefalse
    \hb@xt@\hsize{\hfil\box\@tempboxa\hfil}%
  \fi
  \vskip\belowcaptionskip}
\makeatother

\title{Prior Directions: \\Separating Geometric Identification from\\Behavioral Efficacy in Visual Revision}
\author{\begin{tabular}{c}
Weile Gong\textsuperscript{1}, Zijian Lu\textsuperscript{2}, Mingcai Chen\textsuperscript{3}, Yiping Zuo\textsuperscript{4}, Xin He\textsuperscript{5}, and Weibei Fan\textsuperscript{6}\\
{\normalfont\small Nanjing University of Posts and Telecommunications}\\
{\normalfont\small Nanjing, China}\\
{\normalfont\small \textsuperscript{1} b25011527@njupt.edu.cn; \textsuperscript{2} 18818732360@163.com; \textsuperscript{3} chenmc@njupt.edu.cn;}\\
{\normalfont\small \textsuperscript{4} zuoyiping@njupt.edu.cn; \textsuperscript{5} xhe@njupt.edu.cn; \textsuperscript{6} wbfan@njupt.edu.cn}
\end{tabular}}
\date{}

\begin{document}
\maketitle
\lhead{}
\renewcommand{\headrulewidth}{0pt}

\begin{abstract}
Vision-language models can remain anchored to stale verbal priors even when current visual evidence supports a different answer. We use this controlled revision failure to ask whether the subspace that best captures recurrent prior-induced representation change is also uniquely more behaviorally effective. Across four models, paired prior and reference states define Prior Directions, a compact low-rank geometry that, at rank \(32\), spans only \(0.78\%\) of the state dimensions yet captures \(49.7\%\) to \(64.0\%\) of held-out displacement energy. Matched correspondence improves held-out capture at every tested rank in all four models; across \(20\) permutation seeds, this advantage remains positive in all \(15\) available model–rank settings. Yet the geometric advantage does not translate into a comparable intervention advantage: matched and Pairing-Permuted edits restore \(60/66\) and \(59/66\) clean prior-induced failures, respectively. We trace this discrepancy to the geometry of the realized edits. Pairing-Permuted edits remain strongly aligned with the matched Prior Directions; their aligned component restores \(59/66\) cases, whereas an equally norm-matched orthogonal residual restores only \(20/66\). These results show that the construction that best identifies recurrent representation change need not be uniquely more behaviorally effective, distinguishing global subspace identification from the geometry actually used by an intervention.
\end{abstract}
\begin{center}
\small\textbf{Code:} \url{https://github.com/phare111/prior-directions}
\end{center}

\section{Introduction}

Vision-language models (VLMs) combine current visual input with verbal context from instructions, prior observations, or memory, as in interactive agents \citep{zeng2026mementoguilearningagenticmultimodal}. When the visual state changes, this context can become stale and redirect a correct judgment toward an outdated location. We call this directed revision failure \emph{visual lock-in}.

Benchmarks document language-prior reliance and cross-modal conflict in VLM judgments \citep{lee2025vlind,deng2025words,golovanevsky-etal-2025-pixels}. We isolate this effect in a controlled grounding task derived from ScreenSpot-Pro \citep{li2025screenspotproguigroundingprofessional}. The screenshot, query, and four answers stay fixed while verbal context changes. A stale cue names an outdated location, while irrelevant text controls for added language. Across four models, 69\% to 99\% of semantic-cue losses from initially correct judgments select the prior-consistent location (Figure~\ref{fig:behavior-loss}), making prior-directed errors traceable under fixed visual evidence.

\begin{figure*}[t]
\centering
\includegraphics[width=\textwidth]{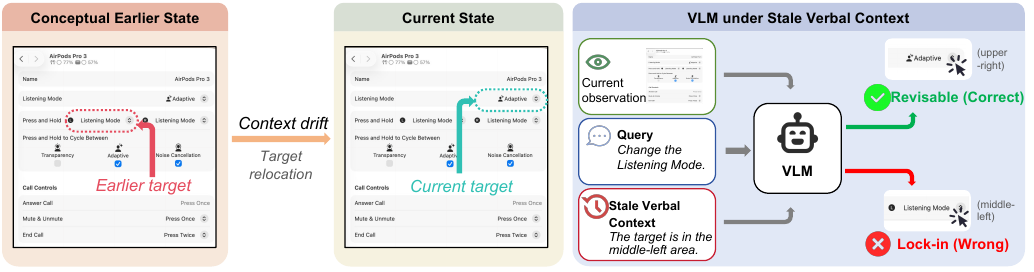}
\caption{Schematic of our controlled visual-revision task with stale verbal context. The conceptual earlier state shows the location described by the verbal prior. Each model receives the current screenshot, query, answer choices, and verbal context. Revisable decisions select the current target, while visual lock-in selects the outdated location.}
\label{fig-teaser}
\end{figure*}

Paired pre-decision states under stale-prior and Irrelevant-Cue conditions define model-specific \emph{Prior Directions}, connecting our analysis to visual task vectors, modality-gap geometry, and activation steering \citep{10.1007/978-3-031-72775-7_15,yan2026crossmodalalignmentmeasuringleveraging,saini2026languageoverwritesvisionoveralignment,wu2025sharp}. The directions generalize to held-out examples. At rank \(32\), they span \(0.78\%\) of the \(4096\)-dimensional state and capture \(49.7\%\) to \(64.0\%\) of held-out prior-induced displacement energy (Figure~\ref{fig:permutation}), compactly describing recurrent change across visual instances.

Prior Directions raise a mechanistic question. We call held-out recurrence \emph{geometric identification} and intervention effects \emph{behavioral efficacy}, two properties studied in activation-space steering and causal editing \citep{liu2025latentspacesteering,jiang2025interpreting,yang2025modularattributionintervention}. They provide complementary evidence. A fitted subspace can capture unseen changes globally while an alternative yields similarly oriented edits on evaluated inputs. Global identification and input-conditioned edit geometry therefore require separate tests.

Following activation-space intervention studies \citep{wu2025sharp,liu2025latentspacesteering}, we compare matched and Pairing-Permuted constructions. Matched correspondence improves held-out capture at every tested rank in all four models, with a positive mean gap in all 15 settings across 20 permutation seeds (Figure~\ref{fig:permutation}). Intervention outcomes are nearly equal, restoring 60/66 and 59/66 clean visual-lock-in cases (Figure~\ref{fig:fig5} and Table~\ref{tab:model-intervention}). Although the fitted subspaces overlap moderately, Pairing-Permuted edits remain aligned with matched Prior Directions on evaluated examples. Their aligned component restores 59/66 cases, compared with 20/66 for an equally norm-matched orthogonal residual. This locates most retained efficacy in the orientation shared by realized edits within each model.

Together, these results make three contributions.
\begin{itemize}
\item \textbf{Prior Directions.} We identify a compact, model-specific geometry of stale-prior-induced revision. At rank \(32\), it spans \(0.78\%\) of state dimensions and captures \(49.7\%\) to \(64.0\%\) of held-out displacement energy across four VLMs.
\item \textbf{Separating geometric identification from behavioral efficacy.} Matched correspondence improves capture in all \(20\) four-model rank comparisons and yields positive mean gaps in all \(15\) repeated-permutation settings across three models. The two constructions restore \(60/66\) and \(59/66\) cases.
\item \textbf{Explaining the separation through realized edit geometry.} The aligned Pairing-Permuted edit restores \(59/66\) cases, compared with \(20/66\) for an equally norm-matched orthogonal residual.
\end{itemize}

\section{Related Work}

\textbf{Representation geometry and intervention.}
Global directions, hallucination subspaces, and selective steering alter vision-language behavior in activation space \citep{liu2025latentspacesteering,yang2025nullumitigatingobjecthallucinations,wu2026revissparselatentsteering,yin2026dynamicmultimodalactivationsteering,Shu2026SteeringVM}. Visual task vectors and modality-gap analyses characterize compact and persistent representation geometry \citep{10.1007/978-3-031-72775-7_15,yan2026crossmodalalignmentmeasuringleveraging}. Recent approaches remove text-aligned components or steer prior-related internal states \citep{saini2026languageoverwritesvisionoveralignment,wu2025sharp}. Probing, corruption, and intervention studies also locate cross-modal interactions associated with hallucination and grounding \citep{golovanevsky2025notice,hua2025conflicting,jiang2025interpreting,yang2025modularattributionintervention}. These studies motivate two empirical objects that merit separate measurement. Held-out recurrence concerns the complete fitted subspace and its behavior on unseen state changes. An intervention concerns the vector generated from that subspace on a particular input. Our diagnostic evaluates each object while varying the prior-to-reference correspondence used in fitting. We then project the realized edit into matched and orthogonal components under equal edit norm. This comparison characterizes how a construction can improve global identification while an alternative construction preserves a behaviorally effective orientation.

\textbf{Cross-modal conflict and stale context.}
Benchmarks document language priors, multimodal inconsistency, and knowledge conflict \citep{lee2025vlind,yan-etal-2025-multimodal,jia2025benchmarkingmultimodalknowledgeconflict,singla2026visionlanguagemodelsguessmeasuring,pmlr-v267-luo25b}. Visual counterfacts and conflicting prompts show that language can redirect decisions despite relevant images \citep{golovanevsky-etal-2025-pixels,deng2025words,rudman2026prompt,ortu-etal-2026-seeing,khayatan2026promptsoverridevisionpromptinduced}. Perception-knowledge interventions further characterize visual defaults and prior override \citep{lietzow2026visiondefault,cheng2026mitigatinghallucinationslargevisionlanguage}. Verbal memory can carry earlier observations into later agent decisions \citep{zeng2026mementoguilearningagenticmultimodal}. Our synthetic stale location creates a controlled conflict with fixed current visual evidence, a shared answer set, and paired states. Initially correct decisions and a designated outdated distractor make the behavioral consequence observable.

\section{A Controlled Testbed for Prior-Driven Visual Revision}
The experiments use a common controlled grounding protocol and model-specific representation spaces.

\paragraph{Models.} We evaluate InternVL3.5-8B, MiniCPM-V-4.5,
Qwen3-VL-8B, and Holo2-8B
\citep{wang2025internvl3_5,minicpm,qwen3vl,hai2025holo2modelfamily}.
All four models enter the primary behavioral, recurrence, and intervention analyses. The repeated-permutation reanalysis uses the saved geometry states from InternVL, Qwen, and Holo.
They span distinct multimodal model families while remaining comparable
in scale.

\begin{table}[!htbp]
\centering
\caption{Cohorts and evaluation roles. Selection for the wording scan is model-specific. Recurrence and intervention use the geometry cohort with distinct fitting protocols.}
\label{tab:cohorts}
\small
\begin{tabular}{@{}lp{0.32\linewidth}p{0.39\linewidth}@{}}
\toprule
Analysis & Cohort & Evaluation rule\\
\midrule
Directed failure & \(1{,}257\) instances per model & Losses conditional on No-Cue correctness\\
Wording scan & Model-specific susceptible cases & Seven ordinal conditions, details in Appendix~\ref{app:force}\\
Recurrence & \(500\) pairs per model & Five folds, each with \(400\) construction and \(100\) held-out pairs\\
Intervention & One \(400/100\) split per model & \(66\) clean failures across four evaluation sets\\
\bottomrule
\end{tabular}
\end{table}

\paragraph{Tasks and Conditions.}
We derive a four-choice controlled conflict protocol from
\textit{ScreenSpot-Pro}~\citep{li2025screenspotproguigroundingprofessional},
using its screenshots, queries, and GUI targets as current evidence.
The source benchmark contains \(1{,}581\) tasks from \(23\) professional
applications across five operating systems.
We construct \(1{,}257\) instances by partitioning each screenshot into
a \(3\times3\) grid and selecting four candidate regions comprising the current
target, an incorrect region designated as its earlier location, and two
distractors.
This transformation preserves the screenshot and query while designating a synthetic earlier location and a matched discrete decision set. Text supplies the synthetic prior about the earlier location.

Each instance is evaluated under the cue set
\(\mathcal{C}=\{0,\mathrm{irr},\mathrm{hyp},\mathrm{fact}\}\).
These are No-Cue, Irrelevant-Cue, Hypothesis-Cue, and Fact-Cue.
Irrelevant-Cue controls for additional text, while the semantic cues
state the same outdated location tentatively or as fact. The screenshot,
query, and candidates remain fixed, yielding \(5{,}028\) evaluations per
model. Greedy decoding returns one of A through D within eight tokens.
Thus cue wording is the only experimental input that varies within an
instance.

\paragraph{Separating discovery from evaluation.}
The analyses use different eligibility rules because they answer different questions (Table~\ref{tab:cohorts}). The initial behavior experiment starts from all constructed instances and measures losses conditional on No-Cue correctness. The wording scan selects previously susceptible cases to characterize responses to verbal certainty. Geometry is fit on a cohort sampled before that behavioral selection. Intervention outcomes are evaluated on clean failures within a disjoint evaluation split. Each denominator therefore corresponds to its evaluation target, comprising baseline-correct judgments, unseen paired changes, or reference-correct prior-induced failures.

\subsection{Directed Failures Validate the Testbed}
We first test whether verbalized priors override correct visual judgments under fixed visual evidence.

\begin{figure}[t]
\centering
\includegraphics[width=\linewidth]{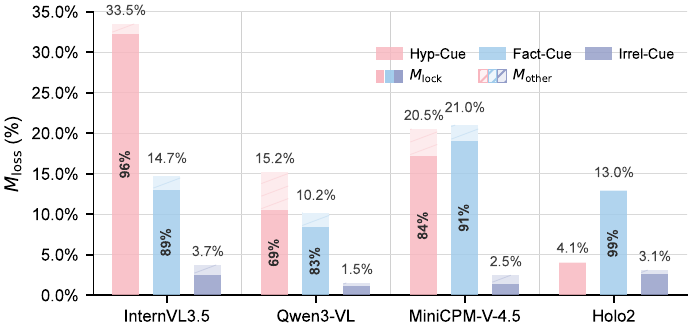}
\caption{Prior cues redirect No-Cue-correct judgments.
Bar height is the full-set lost mass \(M_{\mathrm{loss}}\).
Solid \(M_{\mathrm{lock}}\) and hatched \(M_{\mathrm{other}}\) segments
partition losses into prior-consistent and other errors. In-bar
percentages give DCR, the prior-consistent share of losses.
Gray bars show the Irrelevant-Cue control.}
\label{fig:behavior-loss}
\end{figure}

\paragraph{Behavioral decomposition.}
A case is behaviorally eligible when its No-Cue judgment is correct.
Let \(y_i\) be the visually supported answer, \(y_i^{\mathrm{lock}}\) the prior-consistent distractor, and \(\hat y_{m,i,c}\) the prediction of model \(m\) under cue \(c\).
The Lost-Correct Rate (LCR) and Directed Capture Rate (DCR) are
\begin{equation}
\begin{gathered}
\mathrm{LCR}_m(c)
=\Pr_i(\hat y_{m,i,c}\neq y_i\mid \hat y_{m,i,0}=y_i),\\
\mathrm{DCR}_m(c)
=\Pr_i(\hat y_{m,i,c}=y_i^{\mathrm{lock}}\mid
\hat y_{m,i,0}=y_i,\hat y_{m,i,c}\neq y_i).
\end{gathered}
\end{equation}
LCR measures loss of a correct No-Cue judgment, and DCR measures the
share directed to the prior-consistent option. The full-set masses are
\(M_{m,\mathrm{loss}}(c)=A_{m,0}\mathrm{LCR}_m(c)\) and
\(M_{m,\mathrm{lock}}(c)=M_{m,\mathrm{loss}}(c)\mathrm{DCR}_m(c)\),
where \(A_{m,0}\) is No-Cue accuracy. Figure~\ref{fig:behavior-loss}
partitions this loss mass by destination.
The conditional revision analysis concerns initially correct judgments, and \(A_{m,0}\) supplies their mass in the full test set.

\textbf{Semantic priors induce directed lock-in.}
Figure~\ref{fig:behavior-loss} shows that semantic cues overturn substantially more No-Cue-correct judgments than the Irrelevant-Cue control.
Semantic-cue loss ranges from \(4.1\%\) to \(33.5\%\), while Irrelevant-Cue loss remains between \(1.5\%\) and \(3.7\%\).
Moreover, \(69\%\) to \(99\%\) of semantic-cue losses are directed to the prior-consistent option.
The contrast identifies prior meaning as the organizing factor.

\paragraph{Standardizing the conflict condition.}
We additionally evaluate seven ordinal verbal-certainty conditions,
\(\rho\in\{-3,-2,-1,0,+1,+2,+3\}\), all naming the same incorrect location. The index specifies an ordinal wording condition. We use its highest-certainty endpoint, \(\rho=+3\), as the standardized condition for geometry. Empirically, this endpoint also yields the largest directed lock-in in each model, with model-specific intermediate trajectories. Representative prompts, cohort selection, and the scan appear in Appendix~\ref{app:force}. The scan uses behavior-selected model-specific cohorts. The recurrence analysis uses \(500\) pairs per model sampled before behavioral filtering.

\section{Geometric Identification Through Recurrent Change}
\label{sec:geometry}

A recurring subspace would show that stale-prior influence has structure shared across examples, even when their screenshots and target regions differ. We measure this structure at the \emph{pre-decision position}, the final input token before answer generation. This position provides a common semantic anchor across answer tokenizations. Let \(z^{(\ell)}_{m,i,+3}\) and \(z^{(\ell)}_{m,i,\mathrm{irr}}\) be the hidden states for model \(m\), example \(i\), and layer \(\ell\) under the prior and Irrelevant-Cue reference. Their paired difference is
\begin{equation}
\Delta z^{(\ell)}_{m,i}=z^{(\ell)}_{m,i,+3}-z^{(\ell)}_{m,i,\mathrm{irr}}.
\label{eq:displacement}
\end{equation}
The reference retains additional text without the prior-specific content. We use late layers \(\mathcal L_{\mathrm{late}}=\{24,\ldots,36\}\) throughout. Directions are fit separately for each model and layer. Recurrence refers to transfer across examples within the representation space of each model.

\begin{figure}[t]
\centering
\includegraphics[width=\linewidth]{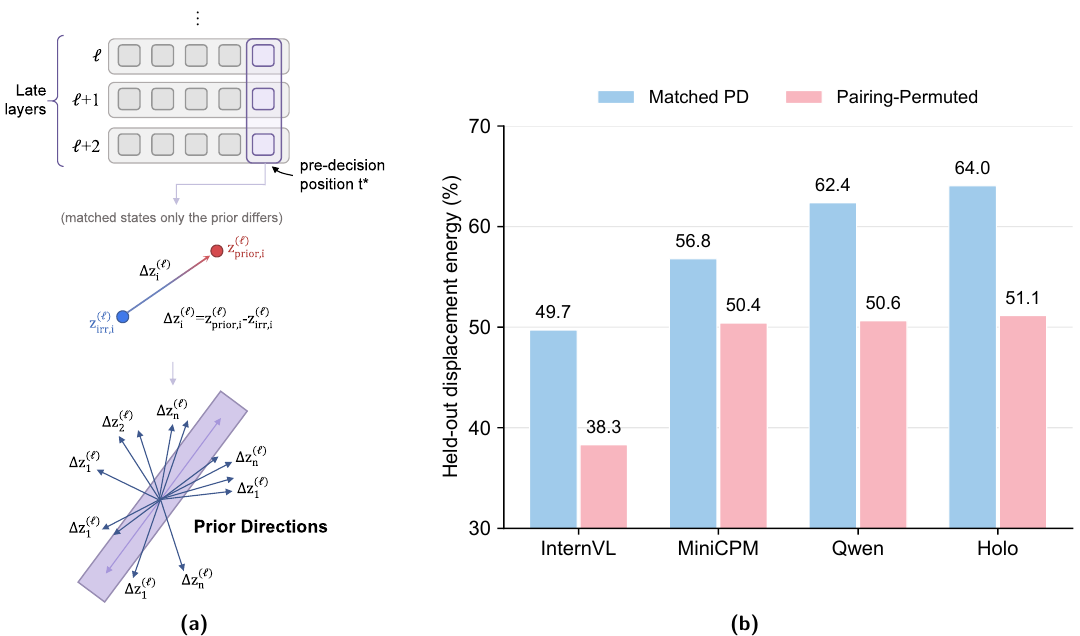}
\caption{Prior Directions as compact recurrent structure in representation change.
(a) Illustrative structural schematic showing how paired pre-decision states define shared low-rank directions across examples.
(b) Held-out displacement energy captured by rank \(32\) matched and Pairing-Permuted directions. Bar labels show the absolute capture percentage for each model, and the isotropic expectation \(k/d=0.78\%\) lies below the displayed range. Matched correspondence exceeds Pairing-Permuted construction in all \(20\) four-model rank settings across the sweep. Across \(20\) permutation seeds, the mean gap is positive in all \(15\) tested settings for three models, as detailed in Appendix~\ref{app:multiseed}.}
\label{fig:permutation}
\end{figure}

\paragraph{Constructing Prior Directions.}
For a construction set \(\mathcal T\), we center its displacements and retain the top \(k\) right singular vectors.
\begin{equation}
\begin{gathered}
\mu^{(\ell)}_{m,\mathcal T}=|\mathcal T|^{-1}\sum_{i\in\mathcal T}\Delta z^{(\ell)}_{m,i},\qquad
\widetilde Z^{(\ell)}_{m,\mathcal T}=[(\Delta z^{(\ell)}_{m,i}-\mu^{(\ell)}_{m,\mathcal T})^\top]_{i\in\mathcal T},\\
\widetilde Z=U\Sigma V^\top,\qquad B_k=[v_1,\ldots,v_k],\qquad P_k=B_kB_k^\top.
\end{gathered}
\label{eq:pd}
\end{equation}
The columns of \(B_k\) are the Prior Directions. Centering removes the condition-wide mean shift from fitting, and SVD summarizes example-dependent variation. Held-out capture below projects the original, uncentered displacement, matching the intervention target. This distinction separates the fitting objective from the evaluation readout. Figure~\ref{fig:permutation}(a) illustrates the paired construction.

\paragraph{Held-out recurrence.}
\label{sec:prior-directions-generalization}
We test recurrence independently of the behavior-selected cohorts by fitting this construction
on a separate geometry cohort \(\mathcal G_m\) of \(500\) controlled
pairs per model at \(\rho=+3\). In fold \(f\), we learn
\(P^{(\ell,k)}_{m,f}\) from \(400\) pairs in
\(\mathcal{T}_{m,f}\subset\mathcal G_m\) and evaluate it on the \(100\)
disjoint pairs in \(\mathcal{H}_{m,f}\), for
\(k\in\{2,4,8,16,32\}\).
Five-fold fitting keeps construction and evaluation disjoint, and the
rank sweep tests recurrence across dimensionalities.
On each held-out fold, \(C^{\mathrm{disp}}\) measures the captured fraction of displacement energy, with \(\epsilon=10^{-12}\) stabilizing all denominators.

\begin{equation}
\begin{gathered}
C^{\mathrm{disp}}_{m,f}(k)
=\underset{\substack{\ell\in\mathcal L_{\mathrm{late}}\\
i\in\mathcal H_{m,f}}}{\operatorname{avg}}
\frac{\|P^{(\ell,k)}_{m,f}\Delta z^{(\ell)}_{m,i}\|_2^2}
{\|\Delta z^{(\ell)}_{m,i}\|_2^2+\epsilon}.
\end{gathered}
\end{equation}
A rank-matched random subspace supplies an isotropic baseline. Pairing-Permuted fits the same construction after shuffling prior-to-reference correspondence within each construction fold. Evaluation retains the original matched held-out differences. Both controls keep rank, folds, and layers fixed.

At \(k=32\), which spans only \(0.78\%\) of the \(4096\)-dimensional
decision state, the matched Prior Directions capture
\(0.497\) to \(0.640\) of held-out
displacement energy, compared with \(0.383\) to \(0.511\) for Pairing-Permuted
(Figure~\ref{fig:permutation}(b)). The matched subspaces exceed random controls across the rank sweep (Appendix~\ref{app:rank}). A random rank-\(32\) subspace has expected capture near \(32/4096=0.0078\), far below the observed matched values. Recurrence is therefore substantial beyond dimensionality alone.

\paragraph{Recurrence across examples.}
The evaluation averages a normalized energy fraction for each example and layer, giving each observation equal weight. The fitted basis summarizes centered variation among construction pairs. Held-out projection measures how much of a fresh, uncentered change falls in that basis. These complementary choices make recurrence an empirical property of unseen changes. The captured fraction reports the geometry available to a subsequent edit, and the intervention tests its behavioral effect.

\section{Matched Correspondence Improves Identification}
\label{sec:correspondence}

The next test isolates the contribution of example correspondence to recurrence. Pairing-Permuted provides a targeted test by preserving the state collections and shuffling their correspondence. For centered construction-state matrices \(A,H\) and a permutation matrix \(\Pi\),
\begin{equation}
\begin{aligned}
X^\top X&=A^\top A+H^\top H-A^\top H-H^\top A,\\
X_\Pi^\top X_\Pi&=A^\top A+H^\top H-A^\top\Pi^\top H-H^\top\Pi A,
\end{aligned}
\end{equation}
where \(X=A-H\) and \(X_\Pi=\Pi A-H\). The marginal terms are unchanged. The cross-covariance terms change. Under a uniform permutation their expectation is zero after centering (Appendix~\ref{sec:prior-directions-mathematical-details}). This identity isolates correspondence as the construction-level difference between the two fitted subspaces.

\paragraph{A consistent gain across ranks and permutation seeds.}
In the original four-model sweep, matched capture exceeds Pairing-Permuted in all \(20\) model and rank comparisons (Appendix~\ref{app:rank}). We repeat the permutation analysis with \(20\) seeds for the three models with saved per-example states, comprising InternVL, Qwen, and Holo. Each model uses the same \(500\) pairs, five folds, split seed \(42\), late layers, and ranks as above. All \(15\) settings retain a positive matched-minus-permutation-mean gap. Every sampled permutation lies below its matched value.

At rank \(32\), the gain is \(0.114\), \(0.117\), and \(0.129\) for InternVL, Qwen, and Holo, respectively. Across settings, permutation SD ranges from \(0.00039\) to \(0.00257\). Even the smallest mean gap, Qwen at rank \(2\), is positive at \(0.008\). The SD measures sensitivity to correspondence shuffling on the fixed cohort. The repeated-seed analysis covers the three models with saved per-example geometry states. Full values and numerical reproduction checks appear in Appendix~\ref{app:multiseed}.

\paragraph{Correspondence contributes at the intervention rank.}
At rank \(16\), the repeated-seed gaps are \(0.101\), \(0.105\), and \(0.124\) for InternVL, Qwen, and Holo. The geometric advantage therefore also holds at the dimensionality used for intervention. Each permuted subspace is refit before evaluation, so this comparison includes the effect of the shuffled covariance on the learned directions. The resulting advantage identifies recurrent structure supplied by matched relations beyond the marginal geometry retained by the control.

Matched correspondence yields consistently higher held-out capture, while the permuted construction retains substantial capture. Behavioral efficacy requires a separate comparison of the edits derived from these constructions on the clean failure cohort.

\section{Behavioral Efficacy Is Preserved Along Shared Edit Orientation}
\label{sec:intervention}

Matched correspondence improves held-out capture, yet the two constructions produce almost the same restoration count. At the evaluated intervention strength, matched removal restores \(60/66\) clean lock-in cases and Pairing-Permuted restores \(59/66\). This separation between identification and efficacy motivates a comparison of the edits each construction actually applies.

For intervention, each \(500\)-pair geometry cohort is split once into
\(400\) construction and \(100\) disjoint evaluation pairs. We fit a
rank-\(16\) subspace on the construction split. Across the four evaluation
sets, \(66\) clean lock-in cases are correct under Irrelevant-Cue and
redirected to \(y_i^{\mathrm{lock}}\) at \(\rho=+3\). This criterion isolates prior-induced failures relative to the same-example reference. The intervention denominator is conditional on Irrelevant-Cue correctness, consistent with the reference used to define the displacement.

\begin{figure}[t]
    \centering
    \includegraphics[width=0.80\linewidth]{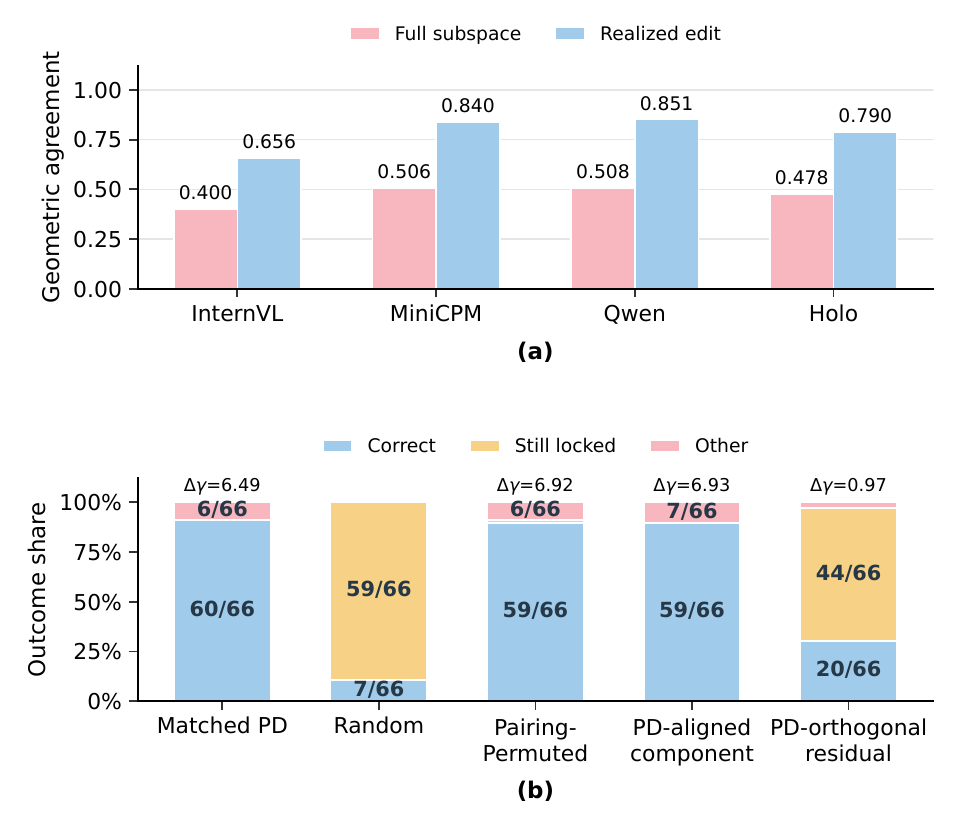}
    \caption{Behavioral efficacy is preserved along shared edit orientation within each model.
    (a) Global projector overlap \(S_{\mathrm{sub}}\) and realized-edit
    alignment \(S_{\mathrm{eff}}\), both relative to matched Prior Directions. (b) Outcomes on \(66\) clean cases
    at \(k=16\) and \(\lambda=0.25\). \(\Delta\gamma\) is the mean
    correct-versus-locked margin change. Random removal is norm-matched to the matched PD edit. Pairing-Permuted and its norm-matched aligned component retain most efficacy, while the
    norm-matched orthogonal residual has a weaker effect. Both components are defined relative to matched Prior Directions.
}
    \label{fig:fig5}
\end{figure}

For each held-out instance, we subtract the component of prior-induced
displacement aligned with the Prior Directions.

\begin{equation}
z_{m,i}^{(\ell),\mathrm{int}}
=
z_{m,i,+3}^{(\ell)}
-
\lambda
P_m^{(\ell,k)}
\Delta z_{m,i}^{(\ell)},
\qquad \lambda=0.25.
\end{equation}
Here \(\Delta z_{m,i}^{(\ell)}\) uses the matched Irrelevant-Cue
reference. Hooks edit only the pre-decision token in
\(\mathcal L_{\mathrm{late}}\), preserving other positions and reference
components. All models and controls share \(\lambda\), rank, layers,
and the clean-case rule. Random and decomposed controls match edit norms per instance and layer to isolate orientation. The subtraction uses reference activations from the same example as a counterfactual diagnostic. All directional comparisons reported here use the common strength \(\lambda=0.25\).

For each edit, we measure answer restoration and the change in correct-versus-locked logit margin,
\(\Delta\gamma_{m,i}=\gamma_{m,i}^{\mathrm{int}}-\gamma_{m,i}^{\mathrm{orig}}\),
where \(\gamma_{m,i}=s_{m,i}(y_i)-s_{m,i}(y_i^{\mathrm{lock}})\) and \(s_{m,i}(y)\) is the option logit.

Matched removal leaves no cases locked, while Pairing-Permuted leaves one locked. Random removal with matched edit norm restores \(7/66\) and leaves \(59/66\) locked, providing a directional control at equal edit magnitude. The similar matched and permuted restoration counts coexist with a clear recurrence gap. To locate the retained effect, we compare
global projector agreement \(S_{\mathrm{sub}}\) for the matched projector \(P\) and Pairing-Permuted projector \(P^{\mathrm{perm}}\) with the fraction of
realized Pairing-Permuted edit energy aligned with the Prior Directions,
\(S_{\mathrm{eff}}\).

\begin{equation}
S_{\mathrm{sub}}
=\frac{1}{k}\operatorname{tr}\!\left(PP^{\mathrm{perm}}\right),
\qquad
S_{\mathrm{eff}}
=\frac{\|PP^{\mathrm{perm}}\Delta z\|_2^2}
{\|P^{\mathrm{perm}}\Delta z\|_2^2+\epsilon}.
\end{equation}
Across models, \(S_{\mathrm{sub}}\) ranges from \(0.400\) to \(0.508\),
with median principal angles of \(42.7^\circ\) to \(53.5^\circ\), while
\(S_{\mathrm{eff}}\) reaches \(0.656\) to \(0.851\)
(Figure~\ref{fig:fig5}(a)). Thus the realized edit retains substantial
alignment despite differences between the full subspaces.
\paragraph{Global geometry and realized edits.}
Global subspace overlap gives every direction equal weight. Realized-edit alignment weights directions by the energy actually present in the edit applied to an example. A Pairing-Permuted subspace can therefore differ from the matched subspace globally while producing an edit that lies largely along matched directions. The principal-angle identity in Appendix~\ref{sec:prior-directions-mathematical-details} formalizes this weighting. The observed alignment calls for a directional intervention that separates the shared component from its orthogonal residual.

We decompose this edit into components aligned with and orthogonal to matched Prior Directions, and match each component to the norm of
\(P\Delta z\).

\begin{equation}
\begin{aligned}
u_{\mathrm{perm}}&=P^{\mathrm{perm}}\Delta z,
&
u_{\parallel}&=\mathcal N(Pu_{\mathrm{perm}}),\\
u_{\perp}&=\mathcal N((I-P)u_{\mathrm{perm}}),
&
\mathcal N(v)&=\frac{\|P\Delta z\|_2}{\|v\|_2+\epsilon}v.
\end{aligned}
\end{equation}
The aligned component restores \(59/66\), leaves none locked, and
raises the margin by \(6.93\) on average. The equally norm-matched
orthogonal residual restores \(20/66\), leaves \(44/66\) locked, and
raises the margin by only \(0.97\). This ordering holds in each model, with the full breakdown in Table~\ref{tab:model-intervention}.
These norm-matched controls locate most retained efficacy in the component aligned with the matched Prior Directions. The orthogonal residual also restores some cases.

\paragraph{Consistent directional effects within each model.}
The aligned component restores \(6/11\) InternVL cases, \(21/23\) MiniCPM cases, \(13/13\) Qwen cases, and \(19/19\) Holo cases. The corresponding orthogonal counts are \(3/11\), \(7/23\), \(2/13\), and \(8/19\). Thus the aggregate restoration contrast is accompanied by the same directional ordering in every model. Matched and permuted removal also have closely spaced restoration counts within each model, differing by one InternVL case and yielding equal counts in the other three models. Each comparison uses a common clean cohort within its model, keeping eligibility fixed across the tested edits.

\begin{table}[!htbp]
\centering
\caption{Intervention outcomes by model. Each entry gives the number correct, still locked, and other, in that order. Aligned and orthogonal refer to components of the Pairing-Permuted edit relative to the matched Prior Directions within each model.}
\label{tab:model-intervention}
\small
\begin{tabular}{@{}lrcccc@{}}
\toprule
Model & Clean cases & Matched & Permuted & Aligned & Orthogonal\\
\midrule
InternVL & 11 & 7 / 0 / 4 & 6 / 1 / 4 & 6 / 0 / 5 & 3 / 8 / 0\\
MiniCPM & 23 & 21 / 0 / 2 & 21 / 0 / 2 & 21 / 0 / 2 & 7 / 14 / 2\\
Qwen & 13 & 13 / 0 / 0 & 13 / 0 / 0 & 13 / 0 / 0 & 2 / 11 / 0\\
Holo & 19 & 19 / 0 / 0 & 19 / 0 / 0 & 19 / 0 / 0 & 8 / 11 / 0\\
\midrule
Total & 66 & 60 / 0 / 6 & 59 / 1 / 6 & 59 / 0 / 7 & 20 / 44 / 2\\
\bottomrule
\end{tabular}
\end{table}

The outcome decomposition further separates releasing a prior-consistent answer from recovering the visual answer. Matched removal and the aligned component leave zero cases locked, with six and seven other errors, respectively. The orthogonal residual leaves \(44\) cases locked. In InternVL, the aligned component restores six cases and moves the remaining five to other answers. In Qwen and Holo, its restored counts cover the entire clean cohort. These model-specific outcomes localize the consistent effect to weakened prior-directed decisions, with the final recovery of the visual answer varying across the evaluated models.

\paragraph{Two geometries in a mechanistic diagnosis.}
Held-out displacement capture measures how well a fitted construction describes recurring change on unseen examples. Effective overlap and component removal measure the realized geometry of an edit on a specific input. Matched pairing improves the first property across the evaluated ranks and models. Pairing-Permuted edits retain much of the second property through directions aligned with the matched subspace, and this component preserves most of their restoration effect. Both levels of geometry are needed to interpret the intervention.

\paragraph{A diagnostic sequence for representation analysis.}
Each control isolates a different relationship between representation geometry and behavior. Rank-matched random subspaces measure capture attributable to dimensionality alone. Shuffling prior and reference correspondence tests what paired construction adds to recurrence while retaining the marginal state collections. Norm-matched random removal then tests the effect of edit magnitude, and the aligned and orthogonal components identify which part of the realized permuted edit carries the observed restoration. Together, these comparisons locate a correspondence-sensitive gain in global identification and an orientation-sensitive behavioral effect. The same sequence can guide interpretation of low-rank interventions in other controlled settings, where the relevant evidence must be measured anew.

\section{Scope and Limitations}

The evidence concerns synthetic stale verbal context under fixed current screenshots, four model families, and standardized English prompts. The single-image protocol isolates revision under controlled conflict. Extending this result to naturally accumulated memory and sequential interaction requires evaluations with observed histories. Recurrence and shared orientation are defined within each model, and broader task transfer remains an open empirical question.

Intervention efficacy is conditional on \(66\) clean failures, the specified reference, late layers, rank \(16\), and strength \(0.25\). The reference supplies same-example counterfactual information, making the edits suitable for mechanistic diagnosis. Their restoration counts characterize this evaluated cohort. Permutation SD measures randomness in construction on fixed examples, with ranks and folds sharing data. The experiments establish a correspondence advantage in recurrence and a dominant aligned component among the tested interventions.

\section{Conclusion}

Stale verbal priors provide a controlled setting in which representation changes are compact and recurrent. Matched correspondence improves the identification of those changes on unseen examples, while Pairing-Permuted edits retain similar restoration efficacy at the tested operating point. Their realized alignment and directional decomposition locate most retained efficacy in an orientation shared within each model. Prior Directions thus expose a useful diagnostic distinction. Geometric identification and behavioral efficacy are complementary tests of a representation-level mechanism, and each requires its own evidence.

\bibliographystyle{references}
\bibliography{prior_directions}

@misc{qwen3vl,
      title={Qwen3-VL Technical Report}, 
      author={Shuai Bai and Yuxuan Cai and Ruizhe Chen and Keqin Chen and Xionghui Chen and Zesen Cheng and Lianghao Deng and Wei Ding and Chang Gao and Chunjiang Ge and Wenbin Ge and Zhifang Guo and Qidong Huang and Jie Huang and Fei Huang and Binyuan Hui and Shutong Jiang and Zhaohai Li and Mingsheng Li and Mei Li and Kaixin Li and Zicheng Lin and Junyang Lin and Xuejing Liu and Jiawei Liu and Chenglong Liu and Yang Liu and Dayiheng Liu and Shixuan Liu and Dunjie Lu and Ruilin Luo and Chenxu Lv and Rui Men and Lingchen Meng and Xuancheng Ren and Xingzhang Ren and Sibo Song and Yuchong Sun and Jun Tang and Jianhong Tu and Jianqiang Wan and Peng Wang and Pengfei Wang and Qiuyue Wang and Yuxuan Wang and Tianbao Xie and Yiheng Xu and Haiyang Xu and Jin Xu and Zhibo Yang and Mingkun Yang and Jianxin Yang and An Yang and Bowen Yu and Fei Zhang and Hang Zhang and Xi Zhang and Bo Zheng and Humen Zhong and Jingren Zhou and Fan Zhou and Jing Zhou and Yuanzhi Zhu and Ke Zhu},
      year={2025},
      eprint={2511.21631},
      archivePrefix={arXiv},
      primaryClass={cs.CV},
      url={https://arxiv.org/abs/2511.21631}, 
}

@misc{minicpm,
      title={MiniCPM-V 4.5: Cooking Efficient MLLMs via Architecture, Data, and Training Recipe}, 
      author={Tianyu Yu and Zefan Wang and Chongyi Wang and Fuwei Huang and Wenshuo Ma and Zhihui He and Tianchi Cai and Weize Chen and Yuxiang Huang and Yuanqian Zhao and Bokai Xu and Junbo Cui and Yingjing Xu and Liqing Ruan and Luoyuan Zhang and Hanyu Liu and Jingkun Tang and Hongyuan Liu and Qining Guo and Wenhao Hu and Bingxiang He and Jie Zhou and Jie Cai and Ji Qi and Zonghao Guo and Chi Chen and Guoyang Zeng and Yuxuan Li and Ganqu Cui and Ning Ding and Xu Han and Yuan Yao and Zhiyuan Liu and Maosong Sun},
      year={2025},
      eprint={2509.18154},
      archivePrefix={arXiv},
      primaryClass={cs.LG},
      url={https://arxiv.org/abs/2509.18154}, 
}

@article{wang2025internvl3_5,
  title={InternVL3.5: Advancing Open-Source Multimodal Models in Versatility, Reasoning, and Efficiency},
  author={Wang, Weiyun and Gao, Zhangwei and Gu, Lixin and Pu, Hengjun and Cui, Long and Wei, Xingguang and Liu, Zhaoyang and Jing, Linglin and Ye, Shenglong and Shao, Jie and others},
  journal={arXiv preprint arXiv:2508.18265},
  year={2025}
}

@misc{hai2025holo2modelfamily,
      title={Holo2 - Open Foundation Models for Navigation and Computer Use Agents}, 
      author={{H Company}},
      year={2025},
      url={https://huggingface.co/collections/Hcompany/holo2}, 
}

@misc{li2025screenspotproguigroundingprofessional,
      title={ScreenSpot-Pro: GUI Grounding for Professional High-Resolution Computer Use}, 
      author={Kaixin Li and Ziyang Meng and Hongzhan Lin and Ziyang Luo and Yuchen Tian and Jing Ma and Zhiyong Huang and Tat-Seng Chua},
      year={2025},
      eprint={2504.07981},
      archivePrefix={arXiv},
      primaryClass={cs.CV},
      url={https://arxiv.org/abs/2504.07981}, 
}

@inproceedings{lee2025vlind,
    title = "{VL}ind-Bench: Measuring Language Priors in Large Vision-Language Models",
    author = "Lee, Kang-il  and
      Kim, Minbeom  and
      Yoon, Seunghyun  and
      Kim, Minsung  and
      Lee, Dongryeol  and
      Koh, Hyukhun  and
      Jung, Kyomin",
    editor = "Chiruzzo, Luis  and
      Ritter, Alan  and
      Wang, Lu",
    booktitle = "Findings of the Association for Computational Linguistics: NAACL 2025",
    month = apr,
    year = "2025",
    address = "Albuquerque, New Mexico",
    publisher = "Association for Computational Linguistics",
    url = "https://aclanthology.org/2025.findings-naacl.231/",
    doi = "10.18653/v1/2025.findings-naacl.231",
    pages = "4129--4144",
    ISBN = "979-8-89176-195-7"
}

@INPROCEEDINGS{deng2025words,
  author={Deng, Ailin and Cao, Tri and Chen, Zhirui and Hooi, Bryan},
  booktitle={2025 IEEE/CVF Conference on Computer Vision and Pattern Recognition (CVPR)}, 
  title={Words or Vision: Do Vision-Language Models Have Blind Faith in Text?}, 
  year={2025},
  volume={},
  number={},
  pages={3867-3876},
  doi={10.1109/CVPR52734.2025.00366}}

@misc{hua2025conflicting,
      title={How Do Vision-Language Models Process Conflicting Information Across Modalities?}, 
      author={Tianze Hua and Tian Yun and Ellie Pavlick},
      year={2025},
      eprint={2507.01790},
      archivePrefix={arXiv},
      primaryClass={cs.CL},
      url={https://arxiv.org/abs/2507.01790}, 
}

@inproceedings{rudman2026prompt,
    title = "Mechanisms of Prompt-Induced Hallucination in Vision{--}Language Models",
    author = "Rudman, William  and
      Golovanevsky, Michal  and
      Arad, Dana  and
      Belinkov, Yonatan  and
      Eickhoff, Carsten  and
      Singh, Ritambhara  and
      Mahowald, Kyle",
    editor = "Liakata, Maria  and
      Moreira, Viviane P.  and
      Zhang, Jiajun  and
      Jurgens, David",
    booktitle = "Proceedings of the 64th Annual Meeting of the {A}ssociation for {C}omputational {L}inguistics (Volume 1: Long Papers)",
    month = jul,
    year = "2026",
    address = "San Diego, California, United States",
    publisher = "Association for Computational Linguistics",
    url = "https://aclanthology.org/2026.acl-long.1941/",
    doi = "10.18653/v1/2026.acl-long.1941",
    pages = "41894--41912",
    ISBN = "979-8-89176-390-6"
}

@inproceedings{golovanevsky2025notice,
    title = "What Do {VLM}s {NOTICE}? A Mechanistic Interpretability Pipeline for {G}aussian-Noise-free Text-Image Corruption and Evaluation",
    author = "Golovanevsky, Michal  and
      Rudman, William  and
      Palit, Vedant  and
      Eickhoff, Carsten  and
      Singh, Ritambhara",
    editor = "Chiruzzo, Luis  and
      Ritter, Alan  and
      Wang, Lu",
    booktitle = "Proceedings of the 2025 Conference of the Nations of the Americas Chapter of the Association for Computational Linguistics: Human Language Technologies (Volume 1: Long Papers)",
    month = apr,
    year = "2025",
    address = "Albuquerque, New Mexico",
    publisher = "Association for Computational Linguistics",
    url = "https://aclanthology.org/2025.naacl-long.571/",
    doi = "10.18653/v1/2025.naacl-long.571",
    pages = "11462--11482",
    ISBN = "979-8-89176-189-6"
}

@inproceedings{jiang2025interpreting,
 author = {Jiang, Nick and Kachinthaya, Anish and Petryk, Suzanne and Gandelsman, Yossi},
 booktitle = {International Conference on Learning Representations},
 editor = {Y. Yue and A. Garg and N. Peng and F. Sha and R. Yu},
 pages = {63582--63605},
 title = {Interpreting and Editing Vision-Language Representations to Mitigate Hallucinations},
 url = {https://proceedings.iclr.cc/paper_files/paper/2025/file/9f14fb9acd243c13c95d4a490d1684ce-Paper-Conference.pdf},
 volume = {2025},
 year = {2025}
}

@misc{lietzow2026visiondefault,
  title={Vision-Default, Prior-Override: Causal Mechanisms of Perception-Knowledge Conflict in Vision-Language Models},
  author={Niclas Lietzow and Danielle Bitterman and Carsten Eickhoff and William Rudman and Michal Golovanevsky},
  year={2026},
  eprint={2606.28273},
  archivePrefix={arXiv},
  primaryClass={cs.CL},
  url={https://arxiv.org/abs/2606.28273}
}

@inproceedings{yan-etal-2025-multimodal,
    title = "Multimodal Inconsistency Reasoning ({MMIR}): A New Benchmark for Multimodal Reasoning Models",
    author = "Yan, Qianqi  and
      Fan, Yue  and
      Li, Hongquan  and
      Jiang, Shan  and
      Zhao, Yang  and
      Guan, Xinze  and
      Kuo, Ching-Chen  and
      Wang, Xin Eric",
    editor = "Che, Wanxiang  and
      Nabende, Joyce  and
      Shutova, Ekaterina  and
      Pilehvar, Mohammad Taher",
    booktitle = "Findings of the Association for Computational Linguistics: ACL 2025",
    month = jul,
    year = "2025",
    address = "Vienna, Austria",
    publisher = "Association for Computational Linguistics",
    url = "https://aclanthology.org/2025.findings-acl.964/",
    doi = "10.18653/v1/2025.findings-acl.964",
    pages = "18829--18845",
    ISBN = "979-8-89176-256-5"
}

@inproceedings{golovanevsky-etal-2025-pixels,
    title = "Pixels Versus Priors: Controlling Knowledge Priors in Vision-Language Models through Visual Counterfacts",
    author = "Golovanevsky, Michal  and
      Rudman, William  and
      Lepori, Michael A.  and
      Bar, Amir  and
      Singh, Ritambhara  and
      Eickhoff, Carsten",
    editor = "Christodoulopoulos, Christos  and
      Chakraborty, Tanmoy  and
      Rose, Carolyn  and
      Peng, Violet",
    booktitle = "Proceedings of the 2025 Conference on Empirical Methods in Natural Language Processing",
    month = nov,
    year = "2025",
    address = "Suzhou, China",
    publisher = "Association for Computational Linguistics",
    url = "https://aclanthology.org/2025.emnlp-main.1262/",
    doi = "10.18653/v1/2025.emnlp-main.1262",
    pages = "24837--24852",
    ISBN = "979-8-89176-332-6"
}

@misc{jia2025benchmarkingmultimodalknowledgeconflict,
      title={Benchmarking Multimodal Knowledge Conflict for Large Multimodal Models}, 
      author={Yifan Jia and Kailin Jiang and Yuyang Liang and Qihan Ren and Yi Xin and Rui Yang and Fenze Feng and Mingcai Chen and Hengyang Lu and Haozhe Wang and Xiaoye Qu and Dongrui Liu and Lizhen Cui and Yuntao Du},
      year={2025},
      eprint={2505.19509},
      archivePrefix={arXiv},
      primaryClass={cs.LG},
      url={https://arxiv.org/abs/2505.19509}, 
}

@misc{khayatan2026promptsoverridevisionpromptinduced,
      title={When Prompts Override Vision: Prompt-Induced Hallucinations in LVLMs}, 
      author={Pegah Khayatan and Jayneel Parekh and Arnaud Dapogny and Mustafa Shukor and Alasdair Newson and Matthieu Cord},
      year={2026},
      eprint={2604.21911},
      archivePrefix={arXiv},
      primaryClass={cs.CV},
      url={https://arxiv.org/abs/2604.21911}, 
}

@inproceedings{yang2025modularattributionintervention,
 author = {Yang, Tianyun and Li, Ziniu and Cao, Juan and Xu, Chang},
 booktitle = {International Conference on Learning Representations},
 editor = {Y. Yue and A. Garg and N. Peng and F. Sha and R. Yu},
 pages = {51546--51568},
 title = {Understanding and Mitigating Hallucination in Large Vision-Language Models via Modular Attribution and Intervention},
 url = {https://proceedings.iclr.cc/paper_files/paper/2025/file/8001c3568152d134d821cd46d4d84768-Paper-Conference.pdf},
 volume = {2025},
 year = {2025}
}

@inproceedings{liu2025latentspacesteering,
 author = {Liu, Sheng and Ye, Haotian and Zou, James Y},
 booktitle = {International Conference on Learning Representations},
 editor = {Y. Yue and A. Garg and N. Peng and F. Sha and R. Yu},
 pages = {72402--72419},
 title = {Reducing Hallucinations in Large Vision-Language Models via Latent Space Steering},
 url = {https://proceedings.iclr.cc/paper_files/paper/2025/file/b4008025c2182bfe16fcc8566ee14d64-Paper-Conference.pdf},
 volume = {2025},
 year = {2025}
}

@misc{yang2025nullumitigatingobjecthallucinations,
      title={Nullu: Mitigating Object Hallucinations in Large Vision-Language Models via HalluSpace Projection}, 
      author={Le Yang and Ziwei Zheng and Boxu Chen and Zhengyu Zhao and Chenhao Lin and Chao Shen},
      year={2025},
      eprint={2412.13817},
      archivePrefix={arXiv},
      primaryClass={cs.CV},
      url={https://arxiv.org/abs/2412.13817}, 
}

@misc{wu2026revissparselatentsteering,
      title={Revis: Sparse Latent Steering to Mitigate Object Hallucination in Large Vision-Language Models}, 
      author={Jialin Wu and Wei Shi and Han Shen and Peigui Qi and Kunsheng Tang and Zhicong Huang and Binghao Wang and Zhou Yang},
      year={2026},
      eprint={2602.11824},
      archivePrefix={arXiv},
      primaryClass={cs.AI},
      url={https://arxiv.org/abs/2602.11824}, 
}

@misc{yin2026dynamicmultimodalactivationsteering,
      title={Dynamic Multimodal Activation Steering for Hallucination Mitigation in Large Vision-Language Models}, 
      author={Jianghao Yin and Qin Chen and Kedi Chen and Jie Zhou and Xingjiao Wu and Liang He},
      year={2026},
      eprint={2602.21704},
      archivePrefix={arXiv},
      primaryClass={cs.CV},
      url={https://arxiv.org/abs/2602.21704}, 
}

@inproceedings{10.1007/978-3-031-72775-7_15,
author = {Hojel, Alberto and Bai, Yutong and Darrell, Trevor and Globerson, Amir and Bar, Amir},
title = {Finding Visual Task Vectors},
year = {2024},
isbn = {978-3-031-72774-0},
publisher = {Springer-Verlag},
address = {Berlin, Heidelberg},
url = {https://doi.org/10.1007/978-3-031-72775-7_15},
doi = {10.1007/978-3-031-72775-7_15},
booktitle = {Computer Vision – ECCV 2024: 18th European Conference, Milan, Italy, September  29–October 4, 2024, Proceedings,  Part XLIII},
pages = {257–273},
numpages = {17},
location = {Milan, Italy}
}

@misc{Shu2026SteeringVM,
  title={Steering Vision-Language Models with Joint Sparse Autoencoders},
  author={Huizhen Shu and Xuying Li and Hongxu Lin and Wenjie Sun and Hui Li},
  year={2026},
  eprint={2606.25657},
  archivePrefix={arXiv},
  primaryClass={cs.CV},
  url={https://arxiv.org/abs/2606.25657}
}

@misc{zeng2026mementoguilearningagenticmultimodal,
      title={MementoGUI: Learning Agentic Multimodal Memory Control for Long-Horizon GUI Agents},
      author={Ziyun Zeng and Hang Hua and Bocheng Zou and Mu Cai and Rogerio Feris and Jiebo Luo},
      year={2026},
      eprint={2605.18652},
      archivePrefix={arXiv},
      primaryClass={cs.CV},
      url={https://arxiv.org/abs/2605.18652},
}

@misc{saini2026languageoverwritesvisionoveralignment,
      title={When Language Overwrites Vision: Over-Alignment and Geometric Debiasing in Vision-Language Models},
      author={Harshvardhan Saini and Samyak Jha and Yiming Tang and Dianbo Liu},
      year={2026},
      eprint={2605.08245},
      archivePrefix={arXiv},
      primaryClass={cs.CV},
      url={https://arxiv.org/abs/2605.08245},
}

@misc{yan2026crossmodalalignmentmeasuringleveraging,
      title={Beyond Cross-Modal Alignment: Measuring and Leveraging Modality Gap in Vision-Language Models},
      author={Hanqi Yan and Xiangxiang Cui and Lu Yin and Jindong Gu and Paul Pu Liang and Yulan He and Yifei Wang},
      year={2026},
      eprint={2502.14888},
      archivePrefix={arXiv},
      primaryClass={cs.CV},
      url={https://arxiv.org/abs/2502.14888},
}

@inproceedings{ortu-etal-2026-seeing,
    title = "When Seeing Overrides Knowing: Disentangling Knowledge Conflicts in Vision-Language Models",
    author = "Ortu, Francesco  and
      Jin, Zhijing  and
      Doimo, Diego  and
      Cazzaniga, Alberto",
    editor = "Liakata, Maria  and
      Moreira, Viviane P.  and
      Zhang, Jiajun  and
      Jurgens, David",
    booktitle = "Proceedings of the 64th Annual Meeting of the {A}ssociation for {C}omputational {L}inguistics (Volume 1: Long Papers)",
    month = jul,
    year = "2026",
    address = "San Diego, California, United States",
    publisher = "Association for Computational Linguistics",
    url = "https://aclanthology.org/2026.acl-long.642/",
    doi = "10.18653/v1/2026.acl-long.642",
    pages = "14109--14130",
    ISBN = "979-8-89176-390-6"
}

@misc{singla2026visionlanguagemodelsguessmeasuring,
      title={Do Vision-Language Models See or Guess? Measuring and Reducing Textual-Prior Reliance with a Phrasing-Controlled Benchmark}, 
      author={Pratham Singla and Shivank Garg and Vihan Singh and Paras Chopra},
      year={2026},
      eprint={2606.10400},
      archivePrefix={arXiv},
      primaryClass={cs.CL},
      url={https://arxiv.org/abs/2606.10400}, 
}

@InProceedings{pmlr-v267-luo25b,
  title = 	 {Probing Visual Language Priors in {VLM}s},
  author =       {Luo, Tiange and Cao, Ang and Lee, Gunhee and Johnson, Justin and Lee, Honglak},
  booktitle = 	 {Proceedings of the 42nd International Conference on Machine Learning},
  pages = 	 {41120--41156},
  year = 	 {2025},
  editor = 	 {Singh, Aarti and Fazel, Maryam and Hsu, Daniel and Lacoste-Julien, Simon and Berkenkamp, Felix and Maharaj, Tegan and Wagstaff, Kiri and Zhu, Jerry},
  volume = 	 {267},
  series = 	 {Proceedings of Machine Learning Research},
  month = 	 {13--19 Jul},
  publisher =    {PMLR},
  url = 	 {https://proceedings.mlr.press/v267/luo25b.html}
}

@misc{cheng2026mitigatinghallucinationslargevisionlanguage,
      title={Mitigating Hallucinations in Large Vision-Language Models via Causal Route Gating}, 
      author={Zhe Cheng and Wenyu Chen and Fode Zhang and Dehuan Shen},
      year={2026},
      eprint={2605.24024},
      archivePrefix={arXiv},
      primaryClass={cs.CV},
      url={https://arxiv.org/abs/2605.24024}, 
}

@inproceedings{wu2025sharp,
  title={{SHARP}: Steering Hallucination in {LVLM}s via Representation Engineering},
  author={Wu, Junfei and Ding, Yue and Liu, Guofan and Xia, Tianze and Huang, Ziyue and Sui, Dianbo and Liu, Qiang and Wu, Shu and Wang, Liang and Tan, Tieniu},
  booktitle={Proceedings of the 2025 Conference on Empirical Methods in Natural Language Processing},
  year={2025},
  pages={14346--14361},
  publisher={Association for Computational Linguistics},
  doi={10.18653/v1/2025.emnlp-main.725},
  url={https://aclanthology.org/2025.emnlp-main.725/}
}

\clearpage
\appendix
\section{Experimental Protocol and Verbal Context}
\label{app:force}

\paragraph{Controlled inputs.}
Each instance fixes the screenshot, current instruction, four candidate regions, correct option, and designated outdated option. The model receives the current screenshot and a prompt specifying the verbal context, instruction, and candidate answers. It returns one letter from A through D using greedy decoding with at most eight generated tokens. The No-Cue control omits the previous-context block. Irrelevant-Cue supplies a generic sentence about unrelated interface controls, application settings, or layout preferences. The fixed experiment seed selects this sentence after excluding messages containing the target-region or prior-region name.

\paragraph{Wording-scan eligibility.}
A pair enters the scan for model \(m\) when its original No-Cue response is correct and at least one semantic cue redirects it to the designated outdated option. Let \(\mathcal I_m\) contain the selected pairs with complete records and \(\Omega_m\) those that remain correct when No-Cue is reevaluated under the common scan wrapper. The sizes \((|\mathcal I_m|,|\Omega_m|)\) are \((417,397)\), \((263,238)\), \((159,139)\), and \((165,148)\) for InternVL, MiniCPM, Qwen, and Holo. Each selected pair produces nine records comprising No-Cue, Irrelevant-Cue, and seven verbal-certainty conditions.

\paragraph{Fixed wording selection.}
Each condition has three semantically matched templates. Representative sentences appear in Table~\ref{tab:force-wording}. The placeholder \texttt{\{region\}} denotes the same incorrect region in every condition for an instance. The template index is fixed by the pair identifier and the ordinal condition before model evaluation,
\[
j(i,\rho)=1+\left(\sum_{c\in\texttt{pair\_id}_i}\operatorname{ord}(c)+\rho\right)\bmod 3.
\]
Thus wording selection is independent of the model output. The index spans \(\rho\in\{-3,-2,-1,0,+1,+2,+3\}\), progressing from uncertain recollection to explicit certainty about the same outdated location.

\paragraph{Behavioral readout.}
The directed lock-in rate over the complete selected cohort is
\[
M_{m,\mathrm{lock}}(\rho)=\frac{1}{|\mathcal I_m|}\sum_{i\in\Omega_m}
\mathbf 1[\hat y_{m,i,\rho}=y_i^{\mathrm{lock}}].
\]
This denominator expresses the baseline-correct fraction multiplied by its conditional redirection rate. Table~\ref{tab:full-force-scan} reports all seven conditions. Intermediate responses vary by model. The highest-certainty endpoint attains the largest directed lock-in in each model and supplies the standardized geometry condition.

\begin{table*}[!htbp]
\centering
\caption{Representative verbal-certainty templates. All conditions name the same outdated region. The complete three-variant template set and original wrapper are preserved in the accompanying source artifact.}
\label{tab:force-wording}
\small
\begin{tabular}{@{}cp{0.87\linewidth}@{}}
\toprule
\(\rho\) & Representative prior sentence\\
\midrule
\(-3\) & With low confidence, I would say the earlier view maybe put the target in the \texttt{\{region\}} region.\\
\(-2\) & A tentative reading of the earlier screen put the target in the \texttt{\{region\}} region.\\
\(-1\) & The earlier screen gave a weak suggestion that the target might have been in the \texttt{\{region\}} region.\\
\(0\) & The earlier note recorded the target location as the \texttt{\{region\}} region.\\
\(+1\) & The earlier interface indicated that the target should be in the \texttt{\{region\}} region.\\
\(+2\) & The earlier interface clearly showed the target located in the \texttt{\{region\}} region.\\
\(+3\) & The earlier screen established that the target was in the \texttt{\{region\}} region.\\
\bottomrule
\end{tabular}

\end{table*}

\begin{table}[t]
\centering
\caption{Directed lock-in across the ordinal wording scan.
Entries report \(M_{\mathrm{lock}}(\rho)\). Intermediate trajectories
are model-specific, while every model reaches its maximum at
\(\rho=+3\).}
\label{tab:full-force-scan}
{\small
\begin{tabular}{@{}c cccc@{}}
\toprule
\(\boldsymbol{\rho}\) & \textbf{InternVL} & \textbf{MiniCPM} &
\textbf{Qwen} & \textbf{Holo} \\
\midrule
\(-3\) & 0.307 & 0.293 & 0.233 & 0.309 \\
\(-2\) & 0.372 & 0.452 & 0.447 & 0.339 \\
\(-1\) & 0.362 & 0.422 & 0.314 & 0.333 \\
\(0\)  & 0.324 & 0.338 & 0.346 & 0.364 \\
\(+1\) & 0.388 & 0.525 & 0.453 & 0.400 \\
\(+2\) & 0.477 & 0.570 & 0.560 & 0.618 \\
\(+3\) & 0.501 & 0.605 & 0.610 & 0.800 \\
\bottomrule
\end{tabular}
}

\end{table}

\section{Recurrence Protocol and Correspondence Controls}
\label{app:rank}

The geometry cohort contains \(500\) pairs per model sampled before behavioral filtering. Five-fold cross-fitting uses split seed \(42\), with \(400\) construction and \(100\) held-out pairs in each fold. Each model and layer has its own centered SVD and projector. Evaluation averages normalized capture over held-out examples and layers \(24\) to \(36\). Table~\ref{tab:full-rank-sweep} gives the original matched and single-permutation results at all five ranks. At rank \(32\), random-subspace capture is \(0.0077,0.0078,0.0078,0.0078\) for InternVL, MiniCPM, Qwen, and Holo.

\begin{table*}[t]
\centering
\caption{Prior-induced displacement recurs along the Prior Directions
across rank.
\(C^{\mathrm{disp}}\) is the fraction of unseen prior-induced
displacement energy captured by the fitted subspace. The matched Prior
Directions exceed Pairing-Permuted in all \(20\) model and rank
comparisons.}
\label{tab:full-rank-sweep}
{\small
\begin{tabular}{@{}c cc cc cc cc@{}}
\toprule
& \multicolumn{2}{c}{\textbf{InternVL}} &
\multicolumn{2}{c}{\textbf{MiniCPM}} &
\multicolumn{2}{c}{\textbf{Qwen}} &
\multicolumn{2}{c}{\textbf{Holo}} \\
\cmidrule(lr){2-3}\cmidrule(lr){4-5}
\cmidrule(lr){6-7}\cmidrule(l){8-9}
\(\boldsymbol{k}\) &
\textbf{Matched} & \textbf{Perm.} &
\textbf{Matched} & \textbf{Perm.} &
\textbf{Matched} & \textbf{Perm.} &
\textbf{Matched} & \textbf{Perm.} \\
\midrule
2  & \textbf{0.107} & 0.046
   & \textbf{0.185} & 0.177
   & \textbf{0.200} & 0.191
   & \textbf{0.199} & 0.171 \\
4  & \textbf{0.189} & 0.093
   & \textbf{0.342} & 0.315
   & \textbf{0.358} & 0.333
   & \textbf{0.327} & 0.284 \\
8  & \textbf{0.277} & 0.164
   & \textbf{0.439} & 0.388
   & \textbf{0.465} & 0.401
   & \textbf{0.437} & 0.358 \\
16 & \textbf{0.371} & 0.270
   & \textbf{0.503} & 0.442
   & \textbf{0.551} & 0.447
   & \textbf{0.540} & 0.415 \\
32 & \textbf{0.497} & 0.383
   & \textbf{0.568} & 0.504
   & \textbf{0.624} & 0.506
   & \textbf{0.640} & 0.511 \\
\bottomrule
\end{tabular}
}

\end{table*}

\subsection{Repeated Permutations}
\label{app:multiseed}

The repeated-permutation analysis uses the saved geometry states of InternVL, Qwen, and Holo. It keeps the original cohorts, folds, layers, and ranks fixed and refits each permuted construction for \(20\) seeds. Evaluation always projects the matched held-out differences. Table~\ref{tab:multiseed} reports the results from the full-precision summary. SD quantifies variability across permutation seeds on these fixed examples. Every sampled permutation lies below matched capture in each of the \(15\) settings.

Recomputed matched capture agrees with the original results to maximum absolute error \(2.44\times10^{-7}\), \(1.12\times10^{-7}\), and \(1.22\times10^{-7}\) for Holo, InternVL, and Qwen. These checks establish numerical continuity with the original computation. MiniCPM contributes its original single-permutation results in Table~\ref{tab:full-rank-sweep}.

\begin{table}[t]
\centering
\caption{Repeated-permutation capture on fixed geometry cohorts. Gap is matched capture minus the permutation mean. SD measures variation over \(20\) permutation seeds.}
\label{tab:multiseed}
\small
\begin{tabular}{@{}lrrrrr@{}}
\toprule
Model & Rank & Matched & Perm. mean & Perm. SD & Gap\\
\midrule
InternVL & 2 & 0.107 & 0.046 & 0.00092 & 0.061 \\
InternVL & 4 & 0.189 & 0.094 & 0.00168 & 0.095 \\
InternVL & 8 & 0.277 & 0.163 & 0.00257 & 0.113 \\
InternVL & 16 & 0.371 & 0.270 & 0.00157 & 0.101 \\
InternVL & 32 & 0.497 & 0.384 & 0.00152 & 0.114 \\
Qwen & 2 & 0.200 & 0.192 & 0.00101 & 0.008 \\
Qwen & 4 & 0.358 & 0.334 & 0.00051 & 0.025 \\
Qwen & 8 & 0.465 & 0.401 & 0.00055 & 0.064 \\
Qwen & 16 & 0.551 & 0.447 & 0.00053 & 0.105 \\
Qwen & 32 & 0.624 & 0.507 & 0.00081 & 0.117 \\
Holo & 2 & 0.199 & 0.171 & 0.00195 & 0.028 \\
Holo & 4 & 0.327 & 0.284 & 0.00039 & 0.043 \\
Holo & 8 & 0.437 & 0.358 & 0.00075 & 0.079 \\
Holo & 16 & 0.540 & 0.416 & 0.00120 & 0.124 \\
Holo & 32 & 0.640 & 0.511 & 0.00139 & 0.129 \\
\bottomrule
\end{tabular}
\end{table}

\section{Intervention Outcomes by Model}
\label{app:intervention}

Intervention uses one seed-\(42\) split with \(400\) construction pairs and \(100\) evaluation pairs per model. A clean failure has a correct Irrelevant-Cue answer and a prior-consistent incorrect answer at \(\rho=+3\). The resulting counts are \(11\), \(23\), \(13\), and \(19\) for InternVL, MiniCPM, Qwen, and Holo. All directional diagnostics use rank \(16\), layers \(24\) to \(36\), and strength \(\lambda=0.25\). Reference activations come from the same instance. Random removal and both decomposed components match the matched-PD edit norm separately for every instance and layer.

Table~\ref{tab:model-intervention} in the main paper separates correct, locked, and other outcomes for each model. The aligned component restores more cases than the orthogonal residual in every model. Summing the model counts reproduces the main-paper totals of \(60\), \(59\), \(59\), and \(20\) restored cases. The reported outcomes are conditional restoration counts under this common diagnostic protocol.

\section{Projection and Intervention Identities}
\label{sec:prior-directions-mathematical-details}

Let \(B=[b_1,\ldots,b_k]\) and \(C=[c_1,\ldots,c_k]\) be orthonormal bases for the matched and Pairing-Permuted subspaces within a model and layer, with projectors \(P=BB^\top\) and \(Q=P^{\mathrm{perm}}=CC^\top\). All readouts and edits depend on these projectors, giving invariance to basis rotations, sign changes, and ordering.

\paragraph{Centered construction objective.}
Let the rows of \(X\) be \((\Delta z_i-\overline{\Delta z})^\top\), with SVD \(X=U\Sigma V^\top\) and descending singular values \(\sigma_j\). Write \(B_k=[v_1,\ldots,v_k]\) and \(P_k=B_kB_k^\top\). For any rank-\(k\) orthogonal projector \(R\),
\[
\begin{aligned}
\|XR\|_F^2&=\operatorname{tr}(RX^\top X)\leq\sum_{j=1}^k\sigma_j^2=\|XP_k\|_F^2,\\
\|X-XR\|_F^2&=\|X\|_F^2-\|XR\|_F^2\geq\|X-XP_k\|_F^2.
\end{aligned}
\]
Thus the fitted Prior Directions minimize squared reconstruction error on centered construction displacements. The empirical recurrence test evaluates the captured fraction of uncentered displacements on disjoint examples.

\paragraph{Permutation control.}
For centered prior and reference matrices \(A,H\in\mathbb R^{n\times d}\), matched and permuted differences are \(X=A-H\) and \(X_\Pi=\Pi A-H\). A uniform permutation preserves both marginal covariance terms. Centering gives \(\mathbb E[\Pi]=n^{-1}\mathbf1\mathbf1^\top\) and
\[
\mathbb E_\Pi[X_\Pi^\top X_\Pi]=A^\top A+H^\top H.
\]
This expectation specifies the covariance effect of disrupting correspondence. The reported capture is computed by fitting and evaluating each sampled permutation separately, retaining the nonlinear dependence of the leading eigenspace on the sampled covariance.

\paragraph{Global and realized overlap.}
Let \(\theta_j\) denote principal angles between the two subspaces. Express the realized permuted edit as \(u=Q\Delta z=\widetilde C\alpha\) in the principal basis of \(Q\). For a nonzero edit,
\[
S_{\mathrm{sub}}=\frac{1}{k}\sum_{j=1}^k\cos^2\theta_j,\qquad
\frac{\|Pu\|_2^2}{\|u\|_2^2}=\sum_{j=1}^k w_j\cos^2\theta_j,\qquad
w_j=\frac{\alpha_j^2}{\sum_r\alpha_r^2}.
\]
Global overlap weights each direction equally. Realized overlap weights by edit energy. The stabilized statistic multiplies the latter expression by \(\|u\|_2^2/(\|u\|_2^2+\epsilon)\). Concentrating edit energy along closely aligned principal directions yields high effective overlap even for moderately overlapping full subspaces.

\paragraph{Norm matching.}
For \(a=\|P\Delta z\|_2\), the control uses \(\mathcal N_\epsilon(v)=av/(\|v\|_2+\epsilon)\), with \(v=Pu\) or \(v=(I-P)u\). This map sends zero to zero. For nonzero \(v\), its norm has relative shortfall \(\epsilon/(\|v\|_2+\epsilon)\) from \(a\). Equal-strength edits therefore control magnitude up to the numerical stabilizer, leaving orientation as the designed difference. The downstream network evaluates these interventions at finite strength, and the restoration counts provide their behavioral readout.

\section{Computational Details}

Original inference and state extraction used a single NVIDIA GeForce RTX 5090 GPU, CUDA, and \texttt{bfloat16} model states. The software stack comprised Python, PyTorch, Transformers, NumPy, and Pillow. Model-specific configurations determine hidden-state extraction and intervention hooks. Generation uses temperature \(0\), disabled sampling, and at most eight output tokens. Each model, condition, and instance combination is evaluated once.

The original configurations use seed \(42\). The repeated-permutation analysis additionally varies the correspondence-shuffling seed over \(20\) runs. It projects saved states using Apple MPS and performs small Gram eigendecompositions on CPU. Across analyses, the numerical stabilizer is \(\epsilon=10^{-12}\). Geometry uses ranks \(2,4,8,16,32\), and the reported intervention comparison uses rank \(16\) and strength \(0.25\).

\end{document}